\pdfoutput=1

\documentclass[11pt]{article}

\usepackage{acl}

\usepackage{times}
\usepackage{latexsym}
\usepackage{multirow}
\usepackage{array}

\usepackage[T1]{fontenc}

\usepackage[utf8]{inputenc}

\usepackage{microtype}

\usepackage{inconsolata}

\usepackage{graphicx}

\usepackage{amsmath}
\usepackage{amssymb}
\usepackage{amsfonts}
\usepackage{algorithm}
\usepackage{algpseudocode}

\usepackage{enumitem}

\usepackage{amsmath,amsfonts,bm}

\def\eqref#1{equation~\ref{#1}}

\def\1{\bm{1}}

\DeclareMathAlphabet{\mathsfit}{\encodingdefault}{\sfdefault}{m}{sl}
\SetMathAlphabet{\mathsfit}{bold}{\encodingdefault}{\sfdefault}{bx}{n}

\def\sV{{\mathbb{V}}}

\def\sX{{\mathbb{X}}}

\newcommand{\R}{\mathbb{R}}

\newcommand{\col}{\texttt}
\newcommand{\agg}{\texttt}
\newcommand{\pattern}{\texttt}

\title{QUIS: Question-guided Insights Generation for Automated Exploratory Data Analysis}

\author{Abhijit Manatkar \\
   IBM Research, India \\
   {\tt abhijitmanatkar@ibm.com} \And
   Ashlesha Akella \\
   IBM Research, India \\
   {\tt ashlesha.akella@ibm.com} \AND
   Parthivi Gupta \thanks{Work done as part of internship at IBM Research, India.} \\
   Indian Institute of Technology Kharagpur \\
   {\tt parthivig@kgpian.iitkgp.ac.in} \And
   Krishnasuri Narayanam \\
   IBM Research, India \\
   {\tt knaraya3@in.ibm.com}
   }

\begin{document}
\maketitle
\begin{abstract}
Discovering meaningful insights from a large dataset, known as Exploratory Data Analysis (EDA), is a challenging task that requires thorough exploration and analysis of the data. Automated Data Exploration (ADE) systems use goal-oriented methods with Large Language Models and Reinforcement Learning towards full automation. However, these methods require human involvement to anticipate goals that may limit insight extraction, while fully automated systems demand significant computational resources and retraining for new datasets. We introduce \textsc{QUIS}, a fully automated EDA system that operates in two stages: insight generation (\textsc{ISGen}) driven by question generation (\textsc{QUGen}). The \textsc{QUGen} module generates questions in iterations, refining them from previous iterations to enhance coverage without human intervention or manually curated examples. The \textsc{ISGen} module analyzes data to produce multiple relevant insights in response to each question, requiring no prior training and enabling \textsc{QUIS} to adapt to new datasets.

\end{abstract}

\section{Introduction}

Exploratory Data Analysis (EDA) is the process of discovering meaningful insights from vast amounts of data, and it is a complex task requiring careful data exploration. There are various EDA techniques to uncover insights by analyzing patterns in the data. Automated Data Exploration (ADE) systems accelerate the EDA process through automation.

ADE literature includes statistics-based \cite{claude2015,ding2019quickinsights,Wang2020DataShotAG,metainsights,insightpilot} and interactive methods \cite{milo2016demosigmod,milo2018next,EDAonSamplesUsingIntents,aaai24text2analysis}, where users explore data through natural language queries or receive suggestions for subsequent actions. Visualization-based techniques \cite{vartak2015seedb,demiralp2017foresight,srinivasan2018augmenting,pacmmod24eda} offer visual insights and allow further queries. However, these methods can become resource-intensive due to extensive user interactions. Goal-oriented ADE approaches, generate insights based on predefined objectives \cite{tang2017extracting,seleznova2020guided,omidvar2022guided,laradji2023capture}. This approach directs the exploration using predefined objectives, such as natural language goals or statistical measures of interestingness. While this reduces user interactions, it may constrain the insights to only those aligned with the predetermined goals.

ADE using reinforcement learning is studied 
\cite{milo2018deep,atenaCIKM19,bar2020automatically,personnaz2021dora,aaai23garg,manatkar2024ilaeda} to achieve full automation. While these 
systems minimize user involvement, they often demand dataset-specific training and substantial computational resources, particularly as the number of features, categorical values, or patterns increases, making the process increasingly challenging.

\subsection{Motivation}

An effective EDA system exercises statistical examination with attention to data semantics, such as analyzing trends in {\em date} and {\em sales price} or examining the impact of {\em weather} on {\em flight delay}. Systems like \cite{demiralp2017foresight,deutch2022fedex,insightpilot,guo2024talk2data} leverage Large Language Models (LLMs) to drive the analysis based on natural language goals. Systems which use LLMs to generate relevant questions based on natural language goals \cite{laradji2023capture}, drive insight discovery based on user queries \cite{wang2022interactive}, and interpret analysis objectives from the user's natural language input to specify desired outcomes \cite{lipman2024linx} have also been proposed.
Guiding EDA through insightful questions enables purposeful exploration, clarifying analysis goals, and deriving actionable insights. In contrast, such a goal-oriented approach \cite{laradji2023capture} may overlook unanticipated critical findings.



\subsection{Our Contributions}
We propose a two-stage ADE system, \textsc{QUIS}, that fully automates the EDA process. In the first stage, \textsc{QUIS} generates questions based solely on the data semantics (dataset information like name, description, column names, and column descriptions) without requiring predefined objectives. In the second stage, \textsc{QUIS} uses statistical analysis to produce insights corresponding to the questions from the first stage. This research contributes to the following advancements

\begin{itemize}[leftmargin=*]
    \item \textbf{Question Generation (\textsc{QUGen})} module generates questions in iterations, where questions generated in previous iterations, along with their reasoning and relevant information, serve as examples for subsequent iterations. This approach helps generate unique questions with broader coverage by providing additional context and guidance to the LLM in each iteration. Our approach eliminates the dependency on manually curated examples and predefined analysis goals.
    \item \textbf{Insight Generation (\textsc{ISGen})} module analyzes the data using statistical patterns and classical search techniques to generate insights in response to the questions from the \textsc{QUGen} module without requiring prior training. For a given question, this module provides multiple relevant insights.
\end{itemize}

QUIS offers notable benefits, including reduced dependency on expert knowledge, enhanced efficiency in the exploration process, the ability to uncover a broader range of insights from the data, and ease of use across various datasets.

\begin{figure*}[t]
    \centering
    \begin{tabular}{c}
      \includegraphics[width=0.74\textwidth]{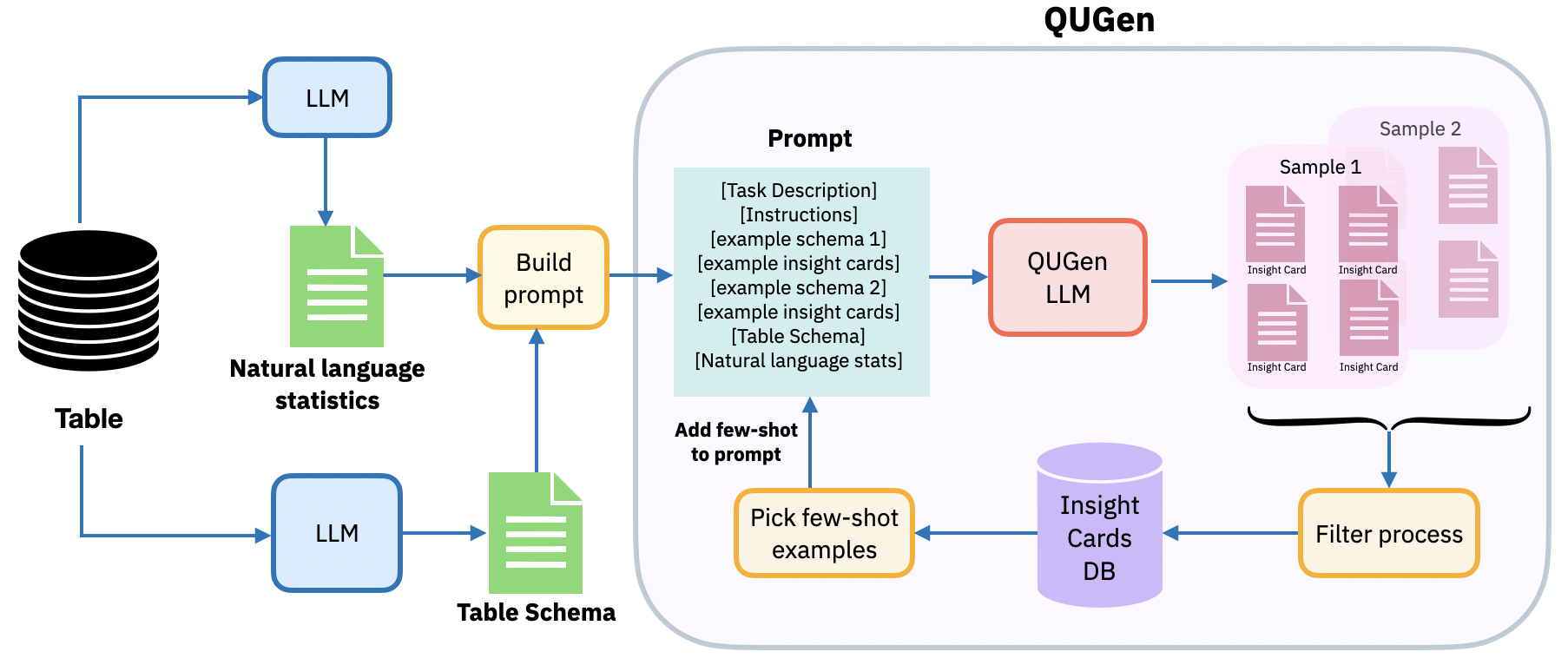}\\
      \hline
      \includegraphics[width=0.74\textwidth]{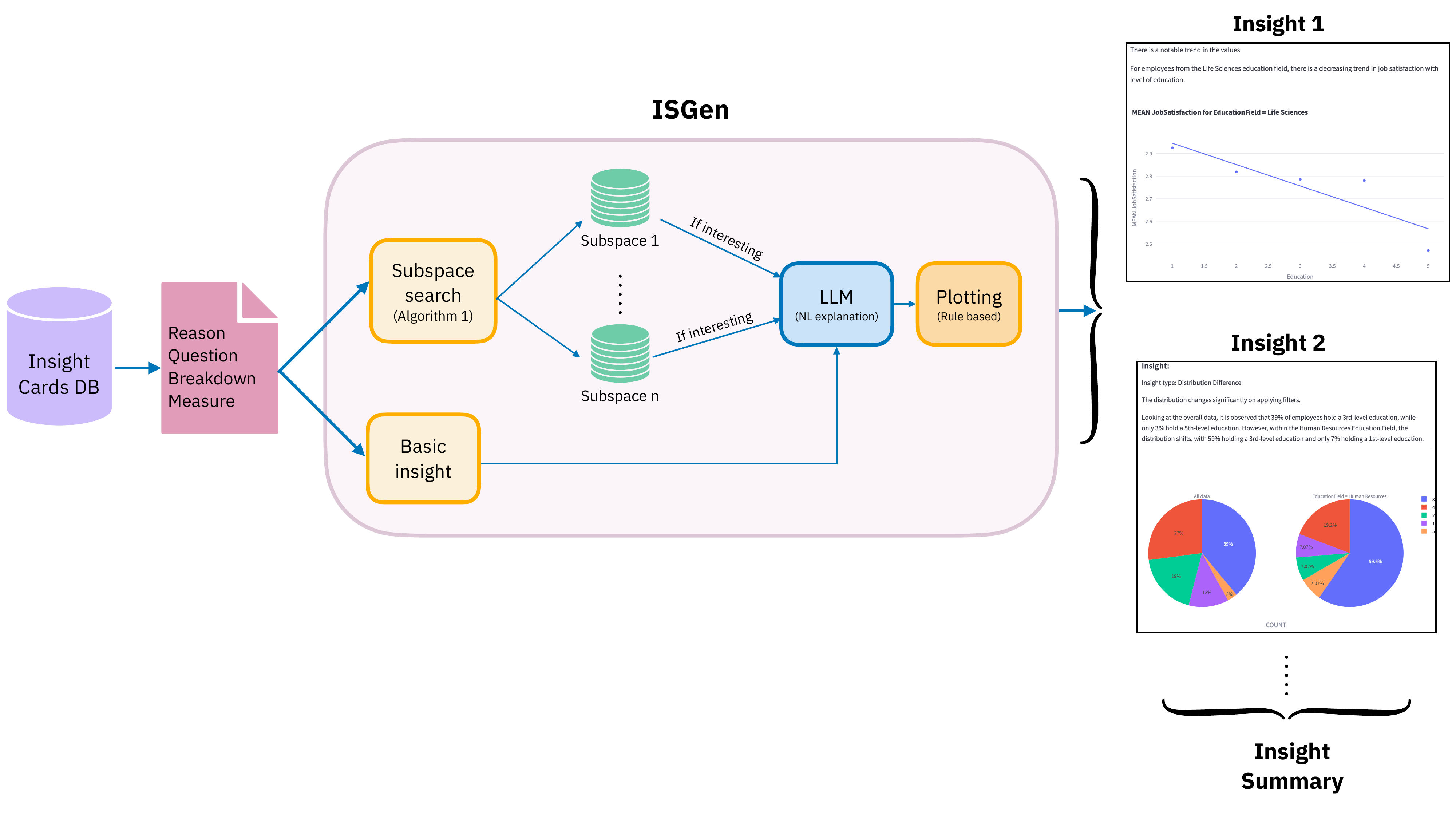}\\
    \end{tabular}
    \caption{The Question Generation (\textsc{QUGen}) module of QUIS system generates questions refined over iterations using data semantics, while the Insight Generation (\textsc{ISGen}) module generates insights (bottom-right) using those questions via statistical analysis. Question is encapsulated inside the Insight Card.}
    \label{fig:arch}
\end{figure*}

\section{Preliminaries}

Although it is challenging to precisely define the notion of an insight due to variations in users' objectives, for this work, we adopt the definition of an insight consistent with previous studies \cite{ding2019quickinsights,insightpilot}. Consider a tabular dataset $D = \{X_1, X_2, \dots X_n\}$ where each $X_i$ is an attribute (column) of the dataset. An insight, denoted by $Insight(B, M, S, P)$, consists of the following:

\begin{enumerate}[leftmargin=*]
    \item \textit{Perspective} - A perspective consists of a tuple $(B, M)$. $B$ represents the \emph{breakdown} attribute, and $M$ is the \emph{measure}, referring to a quantity of interest from the table. Typically, $M$ is of the form  $agg(C)$ where $agg$ (measure function) is an aggregation function , like \texttt{count()}, \texttt{mean()}, \texttt{sum()}, etc., and $C$ (measure column) is a numerical attribute of the dataset. $B$ is the \emph{breakdown} dimension,  a column of interest from the table, for which we want to compare different values of $M$. For each perspective $(B,M)$, we can compute a view $view(D, B, M)$ of the dataset $D$ by grouping on $B$ and calculating the measure $M$ for each group. For example, computing $view(D, \col{Year}, \agg{mean}(\col{Performance}))$ is equivalent to applying the SQL query: \texttt{SELECT Year, AVG(Performance) FROM $D$ GROUP BY Year}. 
    \item \textit{Subspace} - A subspace $S=\bigcup_i\{(X_i, y_{ik})\}$ is a set of filters that determine a subset ($D_S$) of the dataset D. Each $X_i$ is an attribute, and each $y_{ik}$ is a corresponding value of the column $X_i$ of $D$. A tuple $(X_i, y_{ik})$ denotes that the dataset is to be filtered for rows where $D[X_i] = y_{ik}$.
    \item \textit{Pattern} - The pattern $P$ represents the type of insight observed. It belongs to a predefined set of known patterns, such as trends or outliers.
\end{enumerate}

The \textsc{QUIS} system incorporates the following insight types as candidates for our patterns: 
\begin{enumerate}
    \item \textbf{Trend} - An increasing or decreasing trend is seen in a set of values.
    \item \textbf{Outstanding Value} - The largest (or smallest) value in a set of values is significantly larger (or smaller) than all other values in the set.
    \item \textbf{Attribution} - The highest value accounts for a large proportion ($\geq 50\%$) of the total of all values in the set.
    \item \textbf{Distribution Difference} - The distribution of values in a set changes notably from one subspace to another.
\end{enumerate}

As an example, consider the insight given by 
\begin{itemize}[leftmargin=*]
    \item $B = \col{Year}$, $M = \agg{mean}(\col{Performance})$
    \item $S = \{ (\col{Department}, \text{"Sales"}) \}$
    \item $P = \pattern{Trend}$
\end{itemize}

This insight suggests that for the "Sales" department, there has been a trend in the average employee performance over the years. 

By combining a breakdown $B$, a measure $M$, and a subspace $S$, we can compute a unique view of the dataset $D$ by first applying the filters in $S$ on $D$ to arrive at $D_S$, then computing the $view(D_S, B, M)$ as described. Let $\sV(D)$ be the set of all possible views of dataset $D$ that can be computed in this manner. A search for insights involves finding views belonging to $\sV(D)$ for which an insight pattern $P$ is observed. As the size of $\sV(D)$ grows exponentially with the number of columns in $D$, searching for insights by enumerating all possible views in $\sV(D)$ is inefficient. Therefore, it becomes important to limit the search to subspaces that are semantically meaningful and statistically relevant.

\section{Methods}
The EDA process is often guided by the questions that arise from the semantic context and the statistical properties of the dataset. Hence, we propose an approach, \textsc{QUIS} (QUestion-guided InSight generation), that employs a two-stage process (refer to Figure \ref{fig:arch}). The first stage, \textsc{QUGen}, leverages LLMs to formulate questions based on the dataset schema, basic statistics, and iteratively updated in-context examples. The second stage, question-driven insight generation (\textsc{ISGen}), systematically analyzes the tabular data statistics based on the questions to uncover meaningful insights.

\subsection{Question Generation (\textsc{QUGen})}
\begin{figure}[t]
\centering
  \includegraphics[width=0.95\columnwidth]{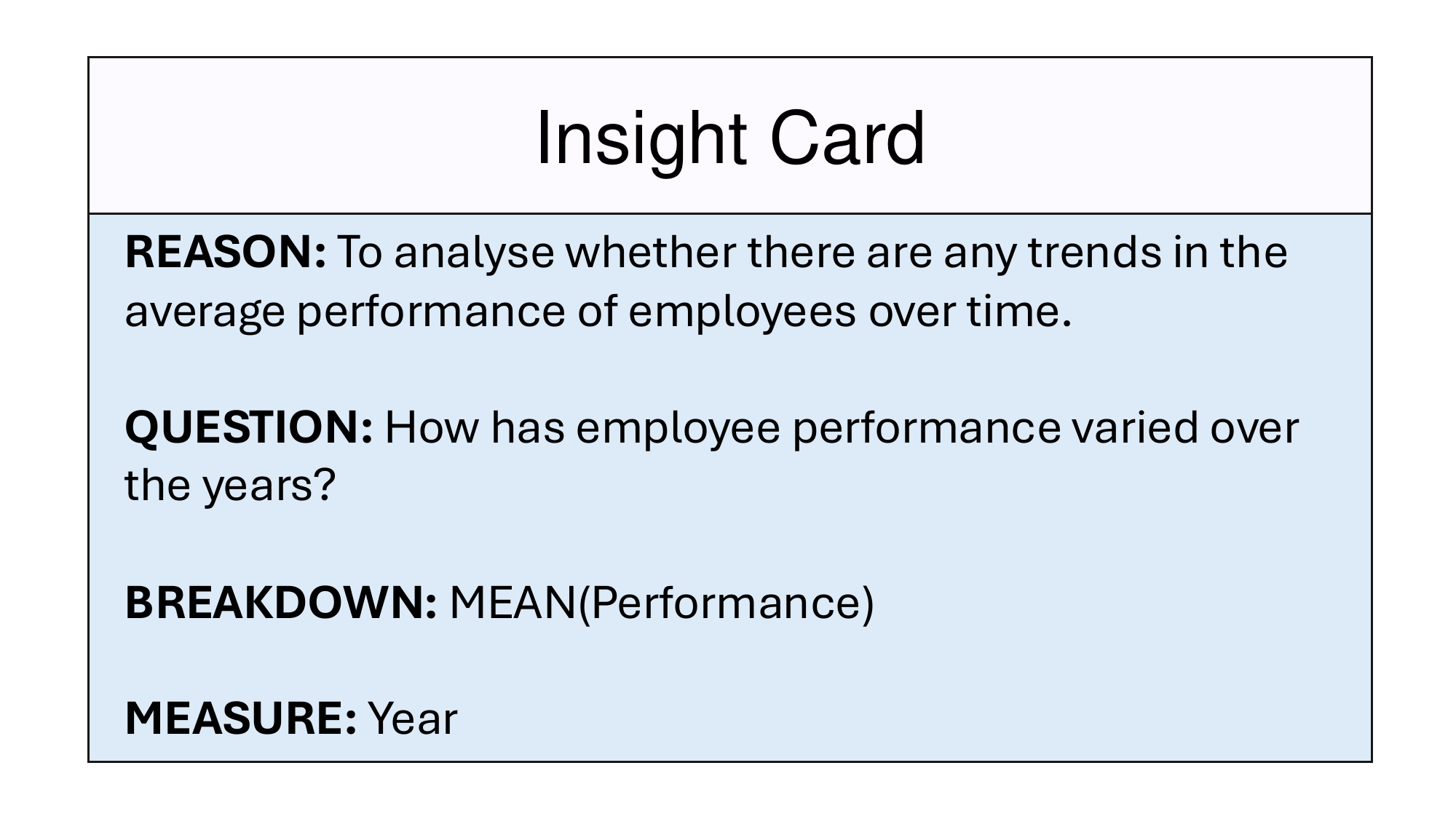}
  \caption{Example Insight Card}
  \label{fig:insight_card}
\end{figure}

Our \textsc{QUIS} framework begins with \textsc{QUGen} producing a set of Insight Cards. Each Insight Card encapsulates relevant information aligning with recent advances in automated EDA \cite{ding2019quickinsights,metainsights}. In particular, an Insight Card (example in Figure \ref{fig:insight_card}) includes four components: \textit{Question}, which is the generated natural language question aimed at guiding data analysis; \textit{Reason}, which explains the rationale behind the generated question to help further analysis; Breakdown $B$, and Measure $M$. The \textit{Reason} is used by \textsc{QUGen} to enhance the coverage, and other components are used by both \textsc{QUGen} and \textsc{ISGen}.

QUGen prompts the language model in a structured way to generate the Breakdown and Measure components, conditioning them on the Reason and Question. This follows the Chain-of-thought prompting approach \cite{chain_of_thought}, where the Reason and Question express the analysis intent behind each insight, ensuring the insights have stronger semantic justification and coherence.

\subsubsection{Input Prompt}
The prompt for \textsc{QUGen} consists of several key components (for details refer to Figure \ref{fig:QUGen_prompt} in Appendix), starting with a high-level description of the data analysis task objective. It then provides detailed instructions for generating an Insight Card by examining the table schema and basic statistics along with a few-shot example table schemas and their sample Insight Cards. Additionally, the prompt includes the schema of the test table and concise natural language descriptions of key statistics summarizing essential information. These statistics are generated by prompting an LLM (for prompt refer Figure \ref{fig:prompt_template} in Appendix) with few-shot examples to generate basic statistical questions, which are transformed into SQL, applied to the dataset, and translated into natural language responses.

\subsubsection{\textsc{QUGen} pipeline}

The \textsc{QUGen} LLM is prompted to generate multiple Insight Cards, as shown in Figure \ref{fig:arch}. The LLM's response is sampled $s$ times with a temperature $t$, with each sample containing $n$ Insight Cards. However, the exact number of Insight Cards per sample may vary slightly due to the fixed output token length.

Each Insight Card undergoes a filtering process: first, cards with questions not semantically relevant to the table schema are removed using semantic similarity computed using the \texttt{all-MiniLM-L6-v2} Sentence Transformers model \cite{reimers-2019-sentence-bert}. Next, duplicate Insight Cards are eliminated based on semantic similarity between pairs of questions. Simple or rudimentary questions are filtered out by converting them to SQL queries and applying them on the dataset; if a query returns only one row, the question is discarded. This ensures that only in-depth questions are retained for comprehensive data analysis.

\textsc{QUGen} is iterative in nature (refer Figure \ref{fig:arch}). It uses subset of Insight Cards generated until the current iteration as in-context examples in the prompt for the next iteration, offering supplementary context and guidance to ensure generation of unique Insight Cards distinct from that of previous iterations. A key advantage of this comprehensive approach by \textsc{QUGen} module is that it eliminates the need for manually providing dataset specific in-context examples, as the Insight Cards generated by the earlier iterations help the LLM understand the dataset context during the subsequent iterations. A collection of Insight Cards accumulated over a certain number (e.g., 10) of iterations are provided as the output by \textsc{QUGen} process.

\subsection{Insight Generation (\textsc{ISGen})}

This module uses classical search techniques and insight scores based on different statistical measures to identify interesting insights from the data.

To determine whether a combination of $B$, $M$, and $S$ reveals a particular pattern $P$, the module uses scoring functions based on data statistics and applies appropriate thresholds. For each insight pattern $P$, a corresponding scoring function $\textsc{ScoreFunc}_{P}: \sV(D) \rightarrow \R$ is defined, along with a threshold value $T_P$. Further details about the scoring function and thresholds for each pattern are provided in Appendix \ref{appendix:scoring_functions}. If a combination of $B$, $M$, and $S$ results in a view $v = view(D_S, B, M)$ such that $\textsc{ScoreFunc}_{P}(v) > T_P$, the insight pattern $P$ is considered to have been observed in $v$.

An Insight Card produced by \textsc{QUGen} module is processed in two stages; first via identifying a basic insight followed by a subspace search for deeper insights as described below.

\subsubsection{Basic Insight}
Extraction of a basic insight helps to depict any meaningful patterns in the relationship between $B$ and $M$ considering the entire dataset without applying any filters. The basic insight is derived from an Insight Card by computing the view $v_0 = view(D, B, M)$. The applicable insight patterns are determined based on the data type of the breakdown $B$ and the measure $M$. For instance, if $B$ is an ordinal column like \col{Year} or \col{Revenue}, then the \pattern{Trend} pattern becomes relevant. Then, scores corresponding to these insight patterns are evaluated. For an insight pattern $P$, if $\textsc{ScoreFunc}_P(v_0) > T_P$, then $Insight(B, M, \phi, P)$ is returned as a basic insight (here $\phi$ is an empty set).

\begin{algorithm}[h]
\caption{Insightful Subspace Search}\label{alg:subspace_search}
\begin{algorithmic}[1]

\Require Dataset $D$, Initial subspace $S_0$, perspective $(B, M)$, language model $LLM$, $\textsc{ScoreFunc}$, \texttt{beam\_width}, \texttt{max\_depth}, \texttt{exp\_factor} 
    \Ensure Top-K subspaces by score $\{ S_1, \dots S_k\}$

    \Function{expand}{$S$}
        \State $\texttt{avlbl\_cols} \gets D.\texttt{cols} - S.\texttt{used\_cols}$
        \Comment{$S.\texttt{used\_cols}$ are the columns used in the filters so far in $S$}
        \State $\texttt{w} \gets \texttt{get\_weights}(\texttt{avlbl\_cols}, LLM)$
        \State $X \gets \texttt{sample(avlbl\_cols, w)}$
        \State $y \gets \texttt{sample}(D[X])$
        \State \Return $S + (X, y)$
    \EndFunction

    \State $\texttt{beam} \gets [(S_0, \Call{ScoreFunc}{S_0})]$
    \For{$\texttt{depth} \in \{ 1, \dots, \texttt{max\_depth} \}$}
        \For{$(S, \texttt{score}) \in \texttt{beam}$}
            \For{$i  \in \{ 1, \dots, \texttt{exp\_factor} \}$}
                \State $S_{new} \gets \Call{expand}{S}$
                \State $\texttt{score} \gets \Call{ScoreFunc}{S_{new}}$
                \State $\texttt{beam}.\texttt{add}((S_{new}, \texttt{score}))$
            \EndFor
        \EndFor
        \State $\texttt{beam} \gets \texttt{top-k}(\texttt{beam},  \texttt{k=beam\_width})$
    \EndFor
    \State \Return \texttt{beam}

\end{algorithmic}
\end{algorithm}

\subsubsection{Subspace Search for Deeper Insights}

Further insights can be generated from an Insight Card by searching for subspaces where the insight patterns are observed. To do so, we carry out a beam search procedure \cite{russell2010artificial} as described in Algorithm \ref{alg:subspace_search}. The search takes an initial subspace $S_0$, a perspective $(B, M)$ and a score function $\textsc{ScoreFunc}_P$ corresponding to insight pattern $P$ as input. A beam of the current best subspaces is maintained. At each step, each subspace $S$ in the beam is expanded to \texttt{exp\_factor} number of subspaces. Each expanded subspace $S_{new}$ is obtained by adding a filter $(X, y)$ to $S$. The selection of $(X, y)$ happens in two steps; selecting the filter column $X$ followed by $y$, the value to filter.

First, an LLM is prompted with $(B,M)$ and an instruction to return candidate filter columns $\sX^{LLM} = \{{X^{LLM}_1 \dots X^{LLM}_k}\}$ that can lead to semantically meaningful insights. $X$ is obtained by sampling from a distribution of available columns (columns of $D$ that have not been used in filters in $S$) with the candidate filter columns $\sX^{LLM}$ having a probability mass of $w_{LLM} \in [0, 1]$ distributed evenly over available columns with the rest of the mass ($1 - w_{LLM}$) distributed over the remaining columns ($D \setminus \sX^{LLM}$). $w_{LLM}$ is decided in such a way to ensure that semantically relevant columns are picked with a high likelihood for filtering while ensuring that other columns also have a chance of being picked.

After picking $X$, we need to pick a value $y$ from $D[X]$. To encourage the selection of values with higher frequency, $y$ is sampled from a distribution over the unique values $\{y_1, \dots y_k\}$ in $D[X]$ where the probability $P(y_i)$ of selecting $y_i$ is given by:
\begin{equation*}
    P(y_i) = \frac{\log(1 + N(y_i))}{\sum_{i}{\log(1 + N(y_i))}}
\end{equation*}
$N(y_i)$ is the frequency $y_i$'s appearance in $D[X]$. 

Each candidate filter $S_{new}$ is evaluated by calculating $\textsc{ScoreFunc}_P(view(D_{S_{new}}, B, M))$ (referred to as $\textsc{ScoreFunc}(S_{new})$ in Algorithm \ref{alg:subspace_search} for conciseness). After a round of expansion and evaluation, the beam is truncated to the \texttt{top-k} (subspace, score) pairs ranked by the score. This process repeats until the maximum desired depth of subspaces, then the final list of subspaces is returned.

The subspaces found in the search procedure are further filtered to only those $S$ for which $\textsc{ScoreFunc}_P(view(D_S, B, M)) > T_P$ to output an insight $Insight(B, M, S, P)$.

\subsubsection{Post Processing}

The post-processing stage of an insight formulates the final insight response, which consists of a natural language description and a corresponding data visualization, as shown in Figure \ref{fig:arch} (\textsc{ISGen}). These components are based on the identified pattern $P$. For each pattern $P$, the natural language response uses a predefined template to clearly communicate the key findings. For details in the plotting conditions for each pattern refer to Appendix \ref{appendix:Plot_per_pattern}.

\section{Experimental Evaluation}

In our study, we evaluated the \textsc{QUIS} pipeline's effectiveness using human assessment and insight scores on three datasets: Sales \cite{onlinesales.dataset}, Adidas Sale \cite{adidas.dataset} and Employee Attrition \cite{attrition.dataset}. Human evaluation focused on the individual insights assessing Relevance, Comprehensibility, and Informativeness (details in Appendix \ref{appendix:eval_criteria}). 
We tested two conditions: 
\begin{enumerate}
    \item \textsc{OnlyStats}, replacing the \textsc{QUGen} module with a purely statistics based card generation module, to assess the autonomous performance of \textsc{ISGen}
    \item \textsc{QUIS}, where both \textsc{QUGen} and \textsc{ISGen} were involved.
\end{enumerate}

Replicating prior work to establish robust baselines \cite{insightpilot,guo2024talk2data,weng2024insightlens} is challenging due to the lack of available code, datasets, and implementation details. Additionally, the differences in insight types and presentation formats across existing approaches make direct comparisons difficult. Therefore, our main focus is on comparing \textsc{QUIS}, against the baseline \textsc{OnlyStats}. For further information about the parameters of the experimental conditions, please refer to Appendix \ref{appendix:parameters}. 

The insights were evaluated by six participants who are well-versed in data analysis, with each insight assessed by three different evaluators. Each criterion - relevance, comprehensibility, and informativeness - was rated on a scale of 1 to 5; where 1 indicated the insight was not relevant, comprehensible, or informative; and 5 indicated the insight was highly relevant, comprehensible, or informative.

\subsection{Human Evaluation}
\begin{figure}[ht]
\centering
  \includegraphics[width=1\columnwidth]{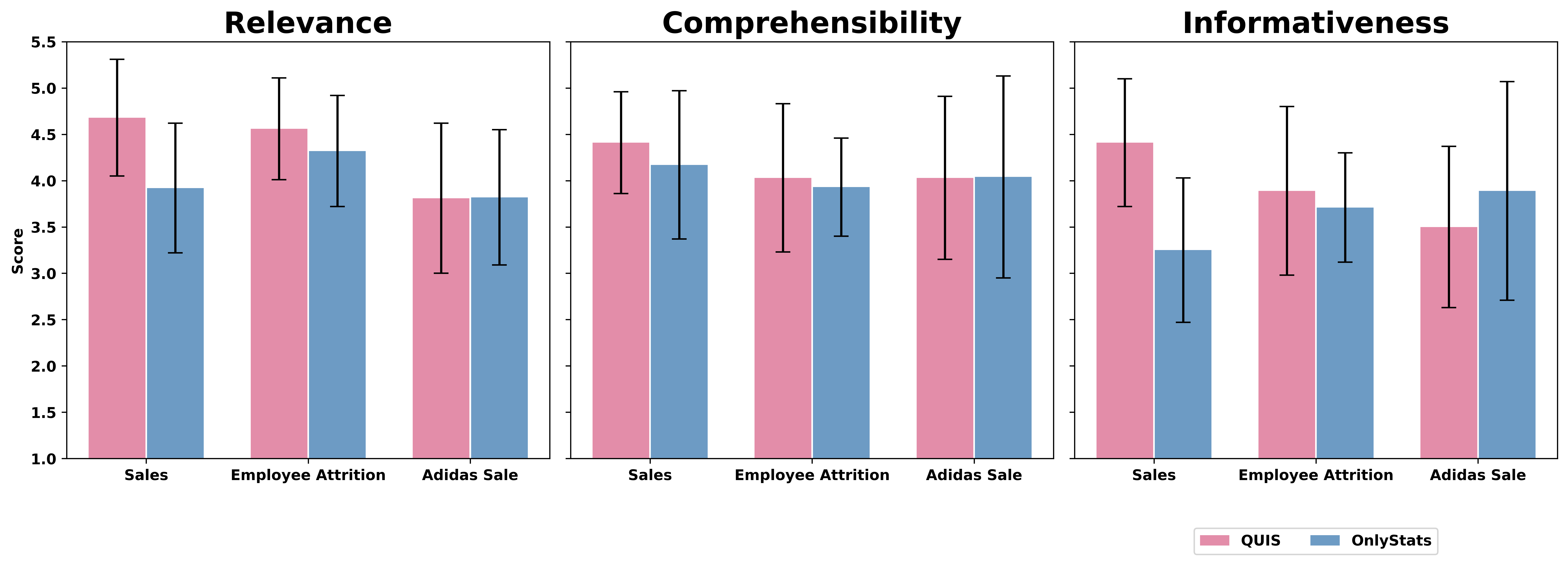}
  \caption{Comparison of Average Human Evaluation Scores for \textsc{QUIS} and \textsc{OnlyStats} across 3 datasets.}
  \label{fig:human_participant}
\end{figure}

The results of the human evaluation in Figure \ref{fig:human_participant} shows that for the Sales and Employee Attrition datasets, QUIS outperformed the \textsc{OnlyStats} baseline in terms of relevance, comprehensibility, and informativeness, suggesting QUIS's overall effectiveness. However, in the Adidas Sales dataset, \textsc{OnlyStats} performed slightly better, likely due to specific characteristics of this dataset which favour a simpler analytical approach. 

\subsection{Insight Score}
\begin{figure}[ht]
\centering
  \includegraphics[width=0.7\columnwidth]{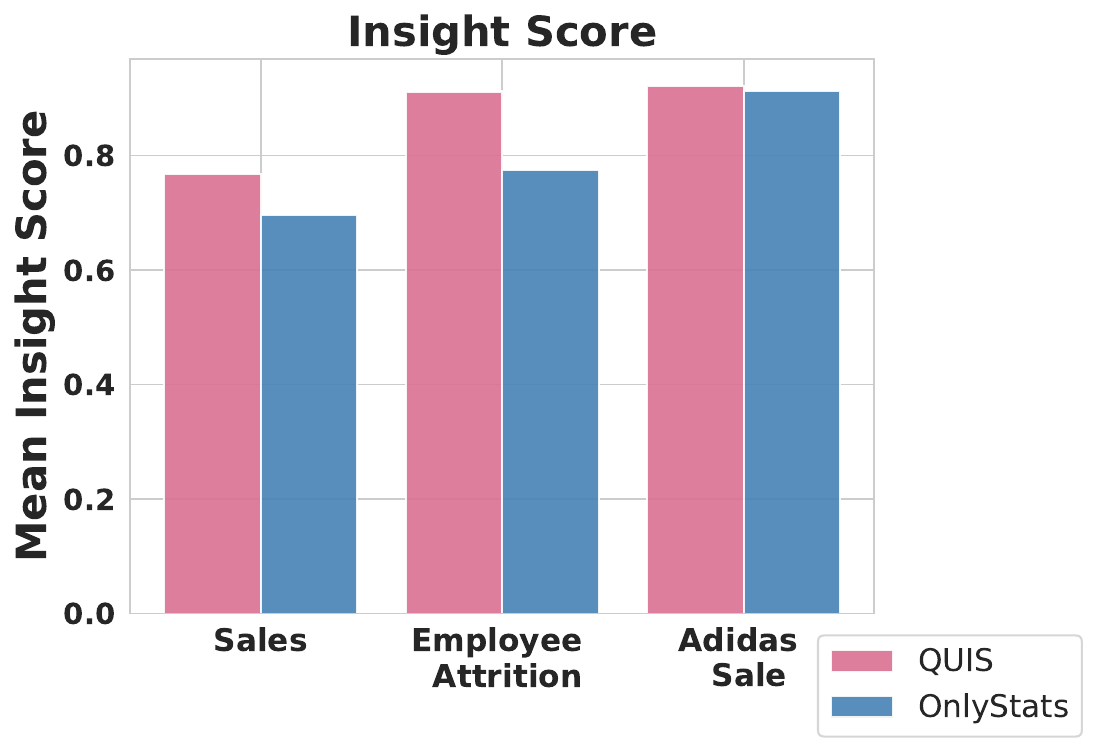}
  \caption{Comparison of Insight score for \textsc{QUIS} and \textsc{OnlyStats}.}
  \label{fig:insight_score}
\end{figure}

We compare the average normalized outputs (in the range $[0, 1]$) of $\textsc{ScoreFunc}$ for all insights returned by the two experimental conditions. The comparison of scores across datasets shows that \textsc{QUIS} consistently outperformed the \textsc{OnlyStats} condition, with higher scores across all datasets as shown in Figure \ref{fig:insight_score}.

\subsection{Diverse Insight Cards}
\begin{figure}[ht]
\centering
  \includegraphics[width=0.9\columnwidth]{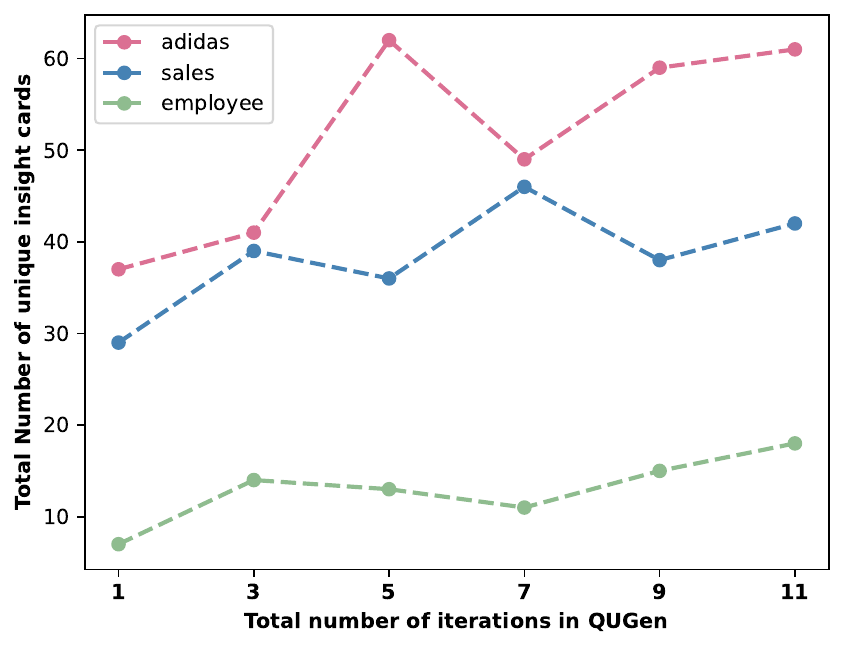}
  \caption{Total number of unique insight cards generated by \textsc{QUIS} under non-iterative (1 iteration) and iterative (up to 11 iterations).}
  \label{fig:diversity_test}
\end{figure}

To assess the effect of the iterative process of  \textsc{QUIS} on Insight Card diversity, we analyzed the number of unique cards generated by \textsc{QUIS} over multiple generations (with varied number of total iterations). We started with 1 iteration and a sampling rate of 20, then progressed to 11 iterations with a sampling rate of 2, keeping the total number of outputs generated by the LLM constant at 20. In the first condition, no few-shot examples were used, while in the last condition, \textsc{QUGen} iterated 10 times, appending the prompt with new few-shot examples sampled from all previous iterations (refer Figure \ref{fig:diversity_test}).

The iterative process produced more diverse Insight Cards, as shown by the rise in the number of unique cards across successive iterations.

\section{Conclusion \& Future Work}

EDA systems often rely on user-generated, goal-oriented questions, which means the quality of the generated insights depends solely on these input questions, introducing potential overhead. To address this limitation, we propose a fully automated EDA system that generates dataset-specific questions automatically and performs insight discovery. This system operates in a data-agnostic manner, requiring no prior training, thereby minimizing the dependency on user input and streamlining the overall insight discovery process.

As a future work, we propose to enhance \textsc{QUGen} to generate questions in chunks where \textsc{ISGen} processes each chunk of questions before \textsc{QUGen} generates the next chunk. This would enable \textsc{QUGen} to use insights and their scores from previous chunks to inform the generation of subsequent chunks. Additionally, we will explore incorporating other types of insights as future work. For example, we aim to include outlier in time-series, anomaly detection, predictive insights and trend reversal to further enhance the variety and depth of insights generated by the \textsc{QUIS} system. 



\bibliography{acl_latex}

\appendix

\section{Scoring Functions for \textsc{ISGen}}
\label{appendix:scoring_functions}

Let $\{v_1, \dots v_k\}$ be the values for view $v \in \sV(D)$ for some dataset $D$.The following scoring functions are defined for measuring the degree to which a particular pattern is seen in $v$.

\begin{enumerate}[leftmargin=*]
    \item Trend - The trend pattern is observed when a sequence of values either increases or decreases monotonically. For quantifying the degree to which the trend pattern is seen, we use the Mann-Kendall Trend Test \cite{mann1945, kendall1975rank}. Specifically we use the implementation in the \texttt{pyMannKendall} package \cite{Hussain2019pyMannKendall}. Let $MK(v)$ return the p-value calculated using the Mann-Kendall test for a $v \in \sV(D)$. Then the score function is given by:
    \begin{equation*}
        \textsc{ScoreFunc}_{\texttt{Trend}}(v) = 1 - MK(v)
    \end{equation*}
    The threshold $T_{\texttt{Trend}}$ is set to $0.95$ so that only views having a p-value $< 0.05$ are returned.
    \item Outstanding Value - The outstanding value pattern is observed when the largest (or most negative) value is much larger (or more negative) than other values. For this pattern, the scoring function calculates the ratio between the largest value in the set and the second largest value in the set. Let $v_{max_{1}}$ and $v_{max_{2}}$ be the two largest (absolute) values in the set. The score is then defined as:
    \begin{equation*}
        \textsc{ScoreFunc}_{\texttt{OV}}(v) = \frac{v_{max_{1}}}{v_{max_{2}}}
    \end{equation*}
    The threshold for this pattern is set to $T_\texttt{OV} = 1.4$
    \item Attribution - The attribution pattern is observed when the top-value in a set of values accounts for more than 50\% of the sum of all values. The score function used for this insight uses the ratio of the largest value to the sum of all values.
    \begin{equation*}
        \textsc{ScoreFunc}_{\texttt{Attr}}(v) = \frac{\max(\{v_1, \dots v_k\})}{\sum_{i}{v_i}}
    \end{equation*}
    As this pattern holds when the highest value is more than 50\% of the total, the threshold is set as $T_\texttt{Attr} = 0.5$.
    \item Distribution Difference - This insight pattern can only be observed when the aggregation in the measure is \texttt{COUNT()}. Let $v^I$ and $v^F$ be the initial and final views. We use the Jensen-Shannon divergence \cite{lin1991jensenshannon} to compare the difference between the two distributions:
    \begin{equation*}
        \resizebox{.9\hsize}{!}{$\textsc{ScoreFunc}_{\texttt{DD}}(v^I, v^F) = JSD(\frac{v^I}{\sum_{i}{v^{I}_i}}||\frac{v^F}{\sum_{i}{v^{F}_i}})$}
    \end{equation*}
    The threshold is set to $T_{\texttt{DD}} = 0.2$.
\end{enumerate}

\section{Plotting per Pattern}
\label{appendix:Plot_per_pattern}
\begin{itemize}
    \item Trend: Scatter plots with trend lines are used to describe the increasing or decreasing nature of the data.
    \item Outstanding Value: Bar charts are used for depicting the difference in the factors.
    \item Attribution: Bar charts are used to show the percentage contribution of different factors
    \item Distribution Difference: Pie charts are used to compare the distributions before and after a condition.
\end{itemize}

\section{Human Evaluation Criteria}
\label{appendix:eval_criteria}

The participants in our user study were asked to rate each generated insight on the following criteria on a scale of 1-5.

\begin{itemize}[leftmargin=*]
    \item Relevance: To what extent the insight is applicable and useful in a given context?
    \item Comprehensibility: To what extent is this insight understandable and easy to follow? 
    \item Informativeness: Does the insight provide substantial information for understanding the data?
\end{itemize}

\section{Experimental Conditions}
\label{appendix:parameters}

\subsection{\textsc{OnlyStats}}

The \textsc{OnlyStats} experimental condition replaces \textsc{QUGen} with a purely statistical method for generating $(B,M)$ pairs as follows. First, a random $B$ is sampled from the list of all eligible columns of the table. This is followed by computing the Kruskal-Wallis test \cite{kruskalwallis} of association between breakdown $B$ and all possible measures $M$ in the table. The Kruskal-Wallis test is a non-parametric variance analysis test, used to determine if two sets of samples come from different distributions. The top 20 pairs of $(B, M)$, ranked according to the strength of association measured by the Kruskal-Wallis test are selected as input to \textsc{ISGen}.

\subsection{\textsc{QUIS}}

For \textsc{QUIS}, the following parameter values were used:

\subsubsection*{\textsc{QUGen}}

\begin{itemize}
    \item LLM: \texttt{Llama-3-70b-instruct} \cite{llama3modelcard}
    \item Sampling temperature $t = 1.1$
    \item Number of samples at each iteration $s = 3$
    \item Number of iterations $n = 10$
    \item Number of in-context examples = 6
\end{itemize}

\subsubsection*{\textsc{ISGen}}

\begin{itemize}
    \item $\texttt{beam\_width} = 100$
    \item $\texttt{exp\_factor} = 100$
    \item $\texttt{max\_depth} = 1$
    \item $w_{LLM} = 0.5$
\end{itemize}

\begin{table*}
\begin{center}
\begin{tabular}{ |>{\centering\arraybackslash}m{4.3cm}|>{\centering\arraybackslash}m{5cm}|>{\centering\arraybackslash}p{5cm}| }

 \hline
 \multicolumn{3}{|c|}{\textbf{Sales Dataset}} \\
 \hline 
 Schema&Sample Questions&Insights\\
 \hline
 \multirow{3}{5cm}{
\vspace{160pt}

Sales ( 

Retailer CHAR\\
\vspace{3pt}
Region CHAR\\
\vspace{3pt}
SalesMethod CHAR\\
\vspace{3pt}
Product CHAR\\
\vspace{3pt}
PricePerUnit INT\\
\vspace{3pt}
UnitsSold INT\\
\vspace{3pt}
TotalSales INT\\
\vspace{3pt}
OperatingProfit INT\\
\vspace{3pt}
OperatingMargin DOUBLE \\
\vspace{3pt}
)}

& \multirow{3}{5cm}{Do products with higher unit prices result in higher total revenue?} &
 \begin{minipage}{5cm}
      \centering
      \vspace{10pt}
      \includegraphics[width=5cm]{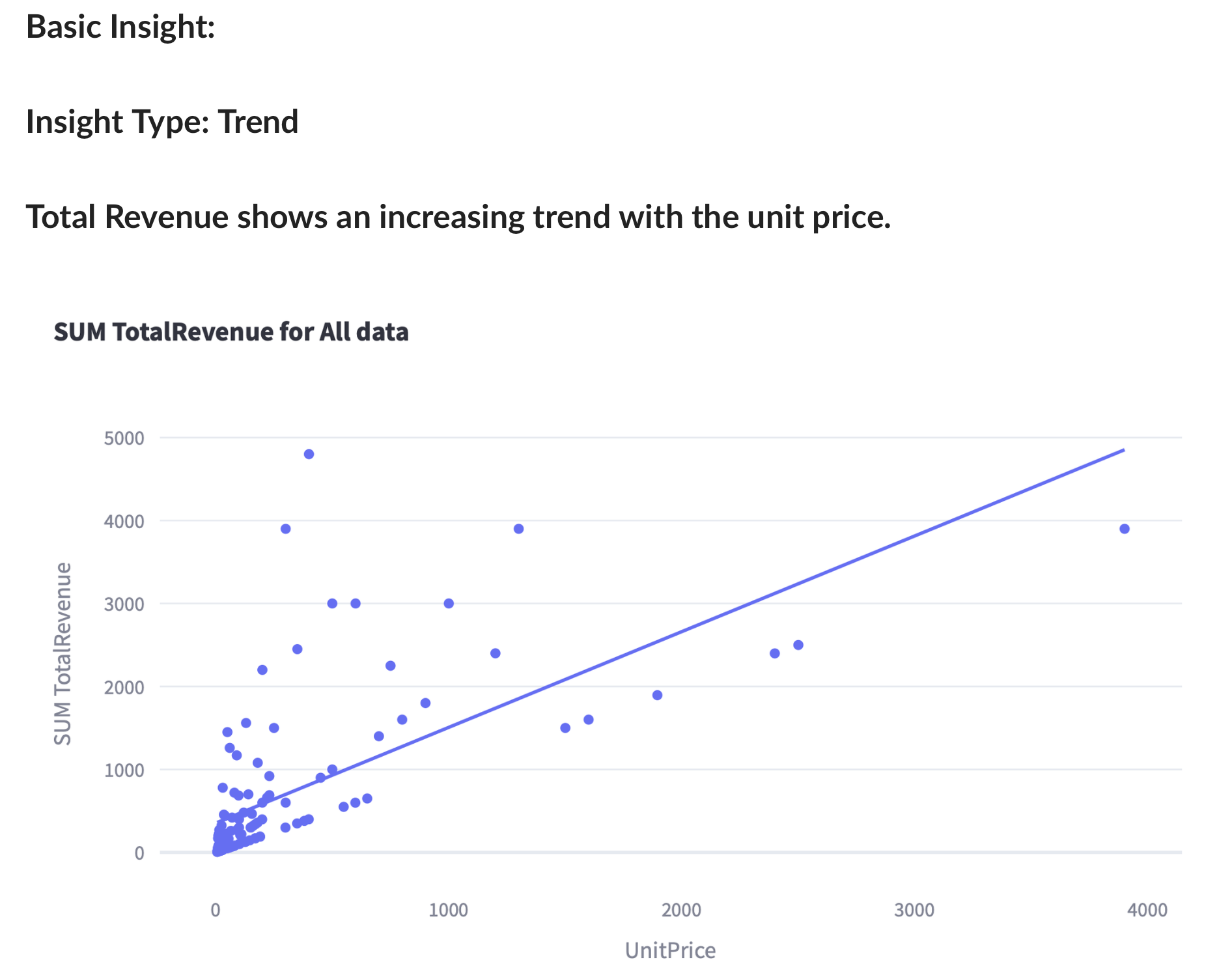} \\
    \end{minipage} \\

 &  &
 \begin{minipage}{5cm}
      \centering
      \vspace{10pt}
      \includegraphics[width=5cm]{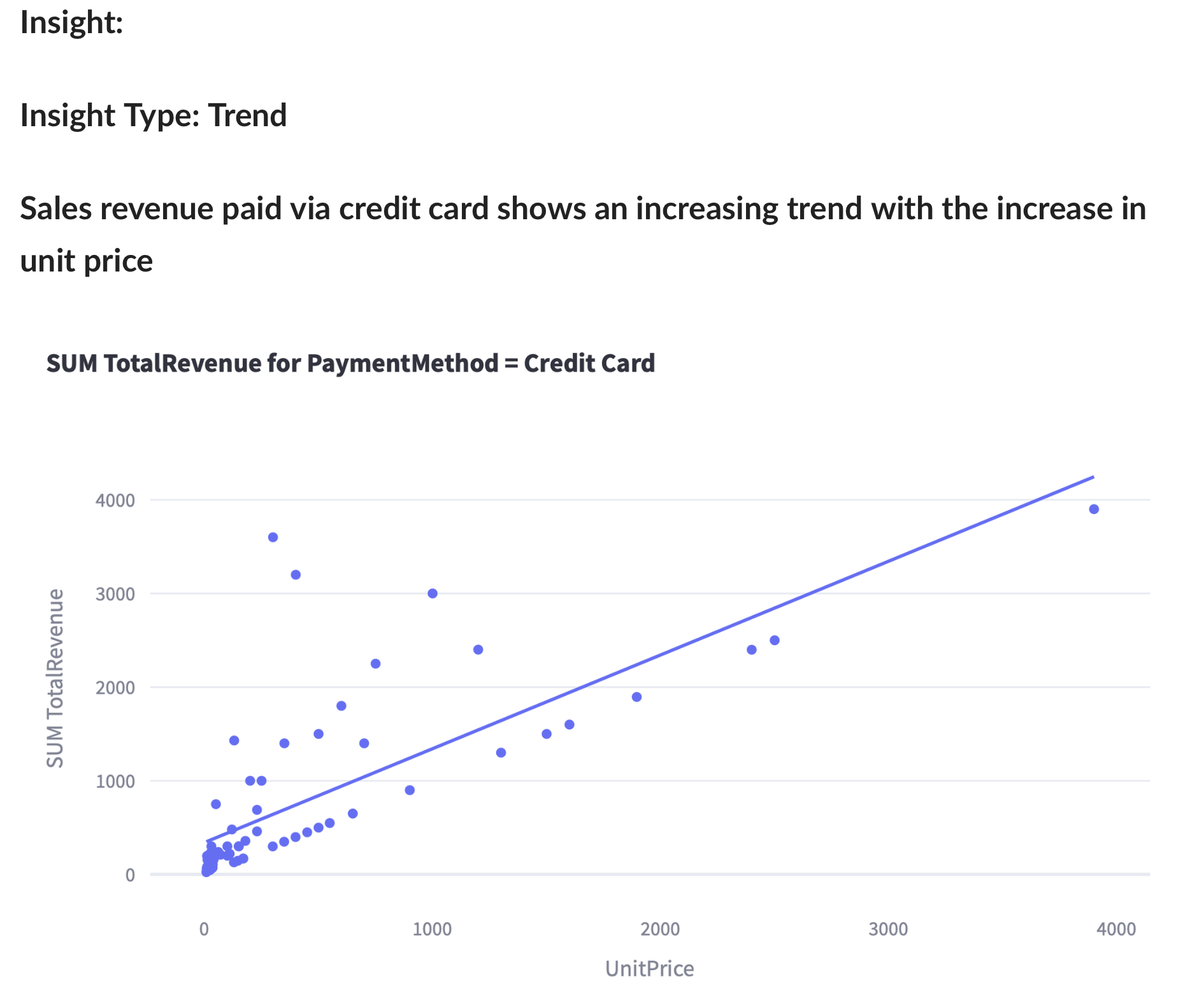} \\
    \end{minipage} \\

 &  &
 \begin{minipage}{5cm}
      \centering
      \vspace{10pt}
      \includegraphics[width=5cm]{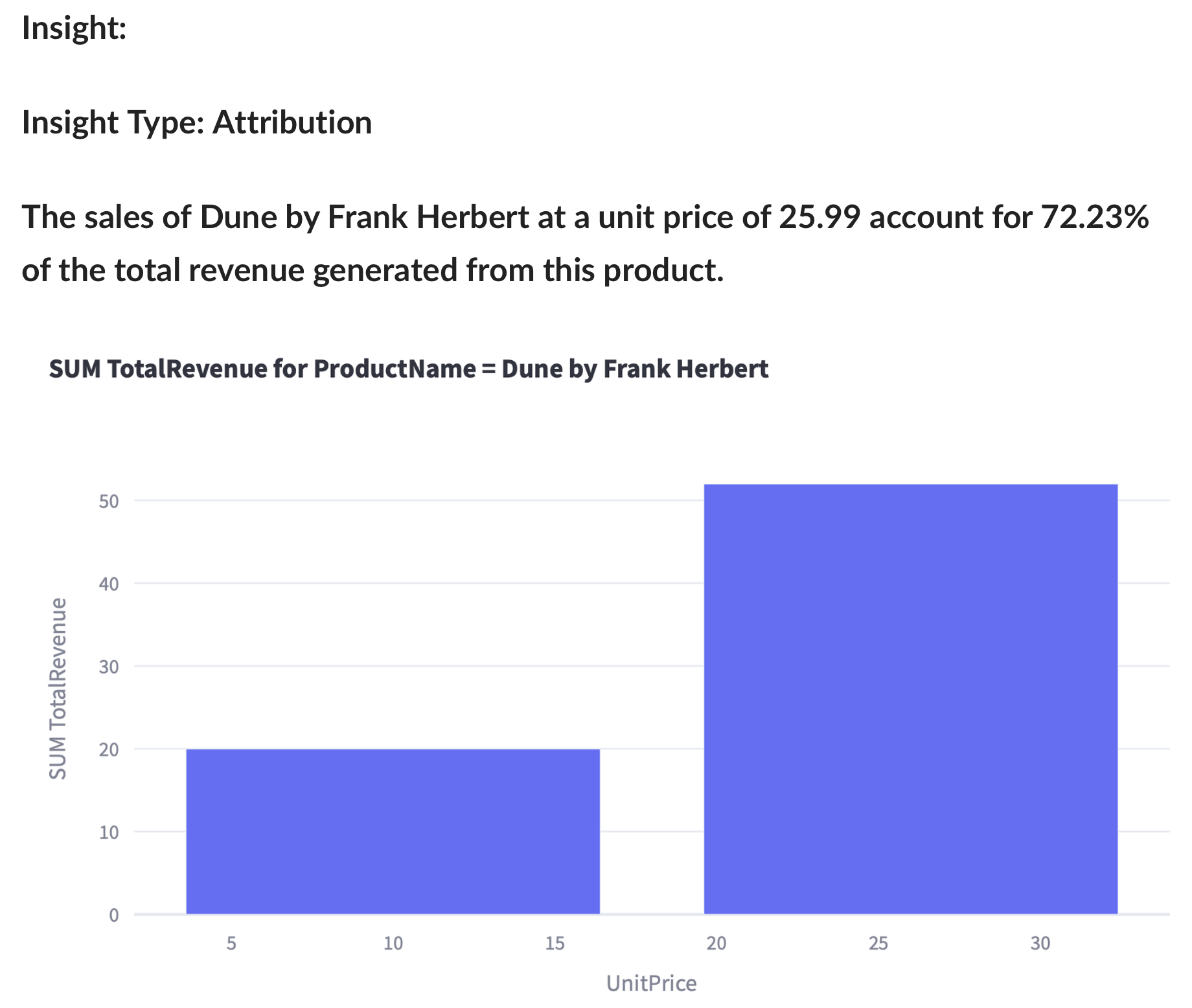} \\
      \vspace{10pt}
    \end{minipage} \\
\cline{2-3}
 
  & \multirow{3}{5cm}{What is the average pricing strategy employed for each product category?} & 
 \begin{minipage}{5cm}
      \centering
      \vspace{10pt}
      \includegraphics[width=5cm]{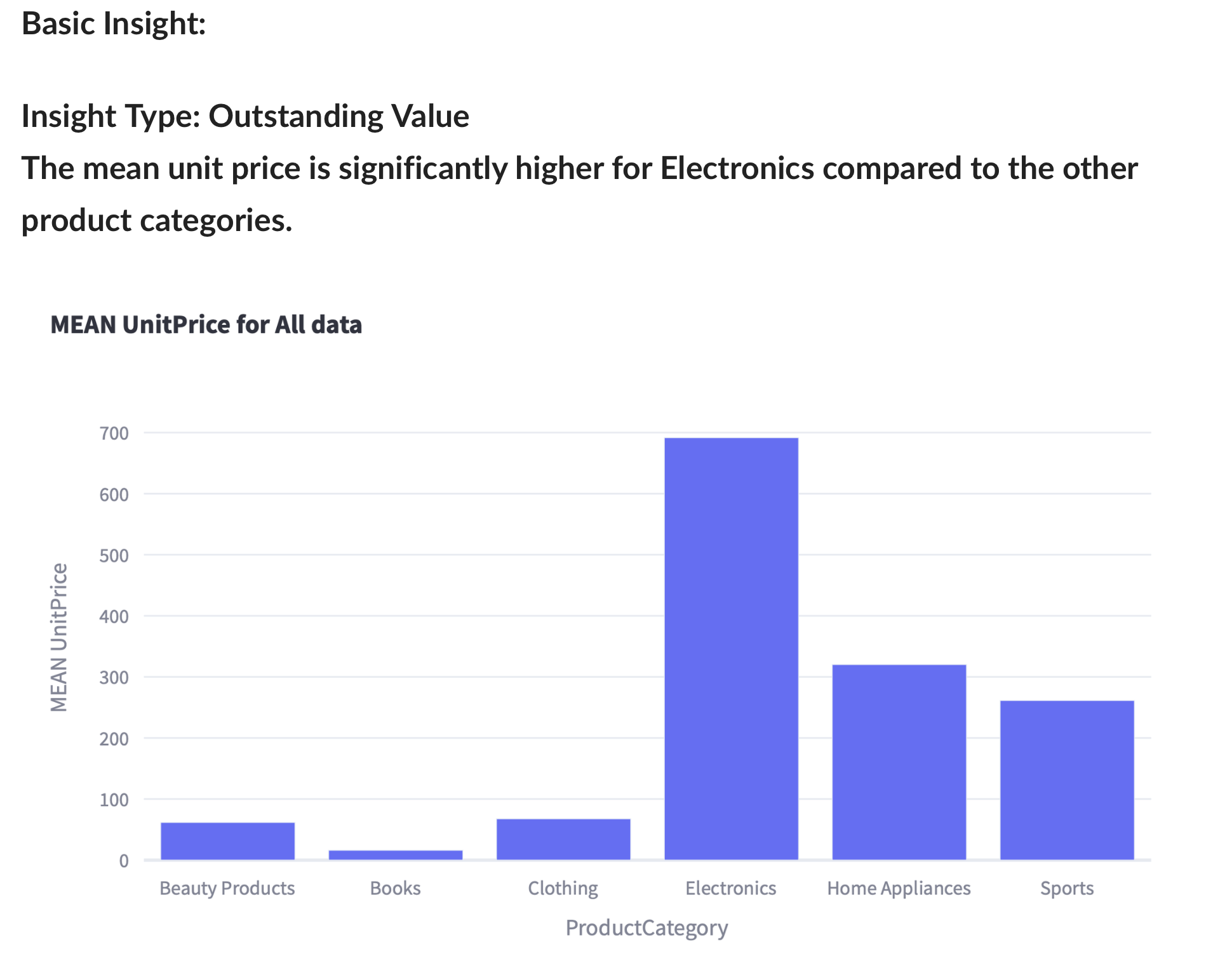} \\
    \end{minipage} \\

 &  &
 \begin{minipage}{5cm}
      \centering
      \vspace{10pt}
      \includegraphics[width=5cm]{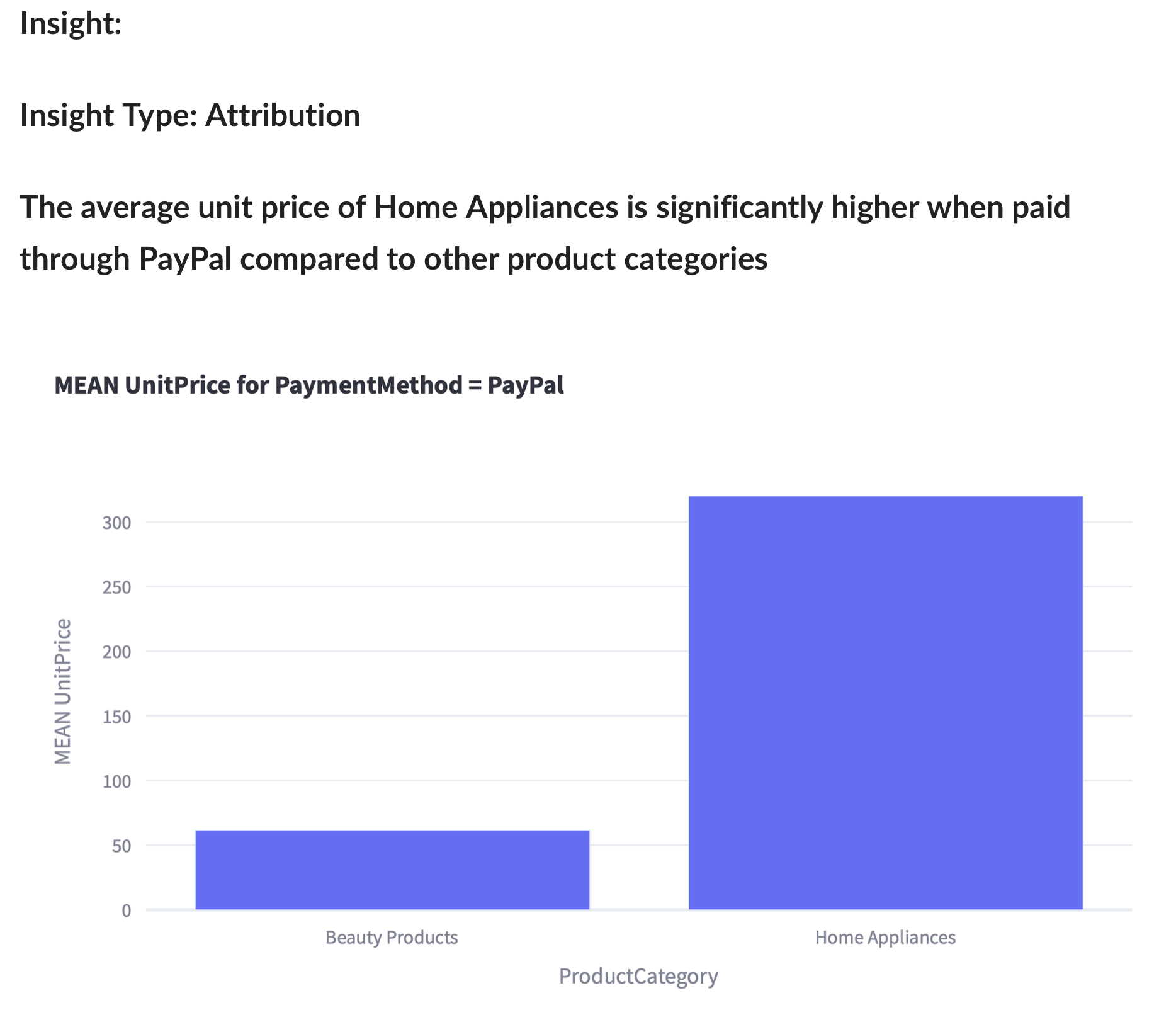} \\
      \vspace{10pt}
    \end{minipage} \\

 \hline

\end{tabular}
\caption{ QUIS Results for Sales Dataset}
\end{center}
\end{table*}


\begin{table*}
\begin{center}
\begin{tabular}{ |>{\centering\arraybackslash}m{4.5cm}|>{\centering\arraybackslash}m{4.5cm}|>{\centering\arraybackslash}p{6cm}| }

 \hline
 \multicolumn{3}{|c|}{\textbf{Employee Attrition Dataset}} \\
 \hline 
 Schema&Sample Questions&Insights\\
 \hline
 \multirow{3}{4.5cm}{

Employee Attrition ( \\
\vspace{3pt}
Age INT \\
\vspace{3pt}
Attrition CHAR \\
\vspace{3pt}
BusinessTravel CHAR \\
\vspace{3pt}
DailyRate INT \\
\vspace{3pt}
Department CHAR \\
\vspace{3pt}
DistanceFromHome INT \\
\vspace{3pt}
Education INT \\
\vspace{3pt}
EducationField CHAR \\
\vspace{3pt}
EmployeeCount INT \\
\vspace{3pt}
EmployeeNumber INT \\
\vspace{3pt}
EnvironmentSatisfaction INT \\
\vspace{3pt}
Gender CHAR \\
\vspace{3pt}
HourlyRate INT \\
\vspace{3pt}
JobInvolvement INT \\
\vspace{3pt}
JobLevel INT \\
\vspace{3pt}
JobRole CHAR \\
\vspace{3pt}
JobSatisfaction INT \\
\vspace{3pt}
MaritalStatus CHAR \\
\vspace{3pt}
MonthlyIncome INT \\
\vspace{3pt}
MonthlyRate INT \\
\vspace{3pt}
NumCompaniesWorked INT \\
\vspace{3pt}
Over18 CHAR \\
\vspace{3pt}
OverTime CHAR \\
\vspace{3pt}
PercentSalaryHike INT \\
\vspace{3pt}
PerformanceRating INT \\
\vspace{3pt}
RelationshipSatisfaction INT \\
\vspace{3pt}
StandardHours INT
)
}
 
  & \multirow{3}{4.5cm}{What is the relationship between employees' Education levels and their Attrition rates?} & 
 \begin{minipage}{6cm}
      \centering
      \vspace{10pt}
      \includegraphics[width=6cm]{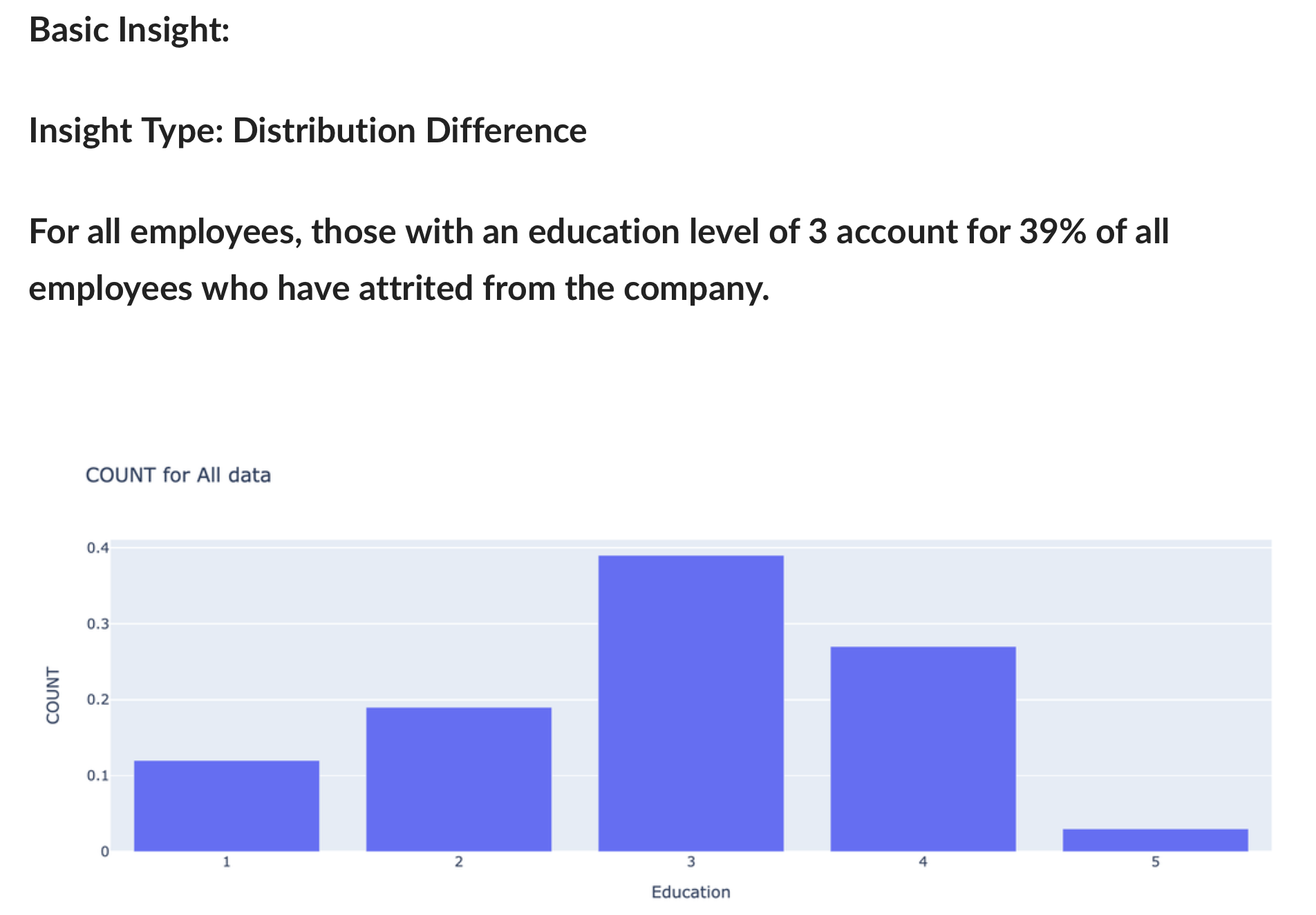} \\
    \end{minipage} \\

 &  &
 \begin{minipage}{6cm}
      \centering
      \vspace{10pt}
      \includegraphics[width=6cm]{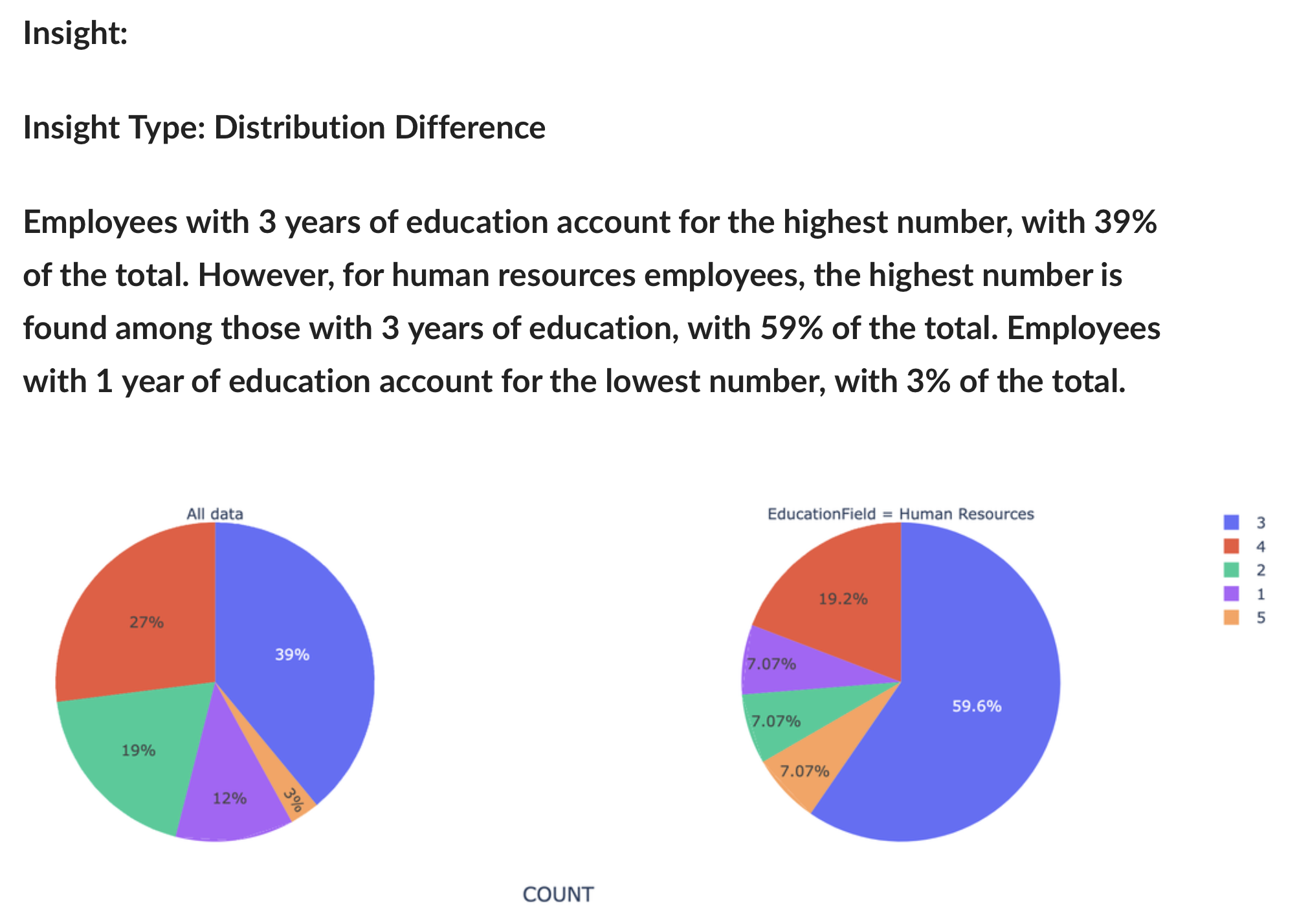} \\
    \end{minipage} \\

 &  &
 \begin{minipage}{6cm}
      \centering
      \vspace{10pt}
      \includegraphics[width=6cm]{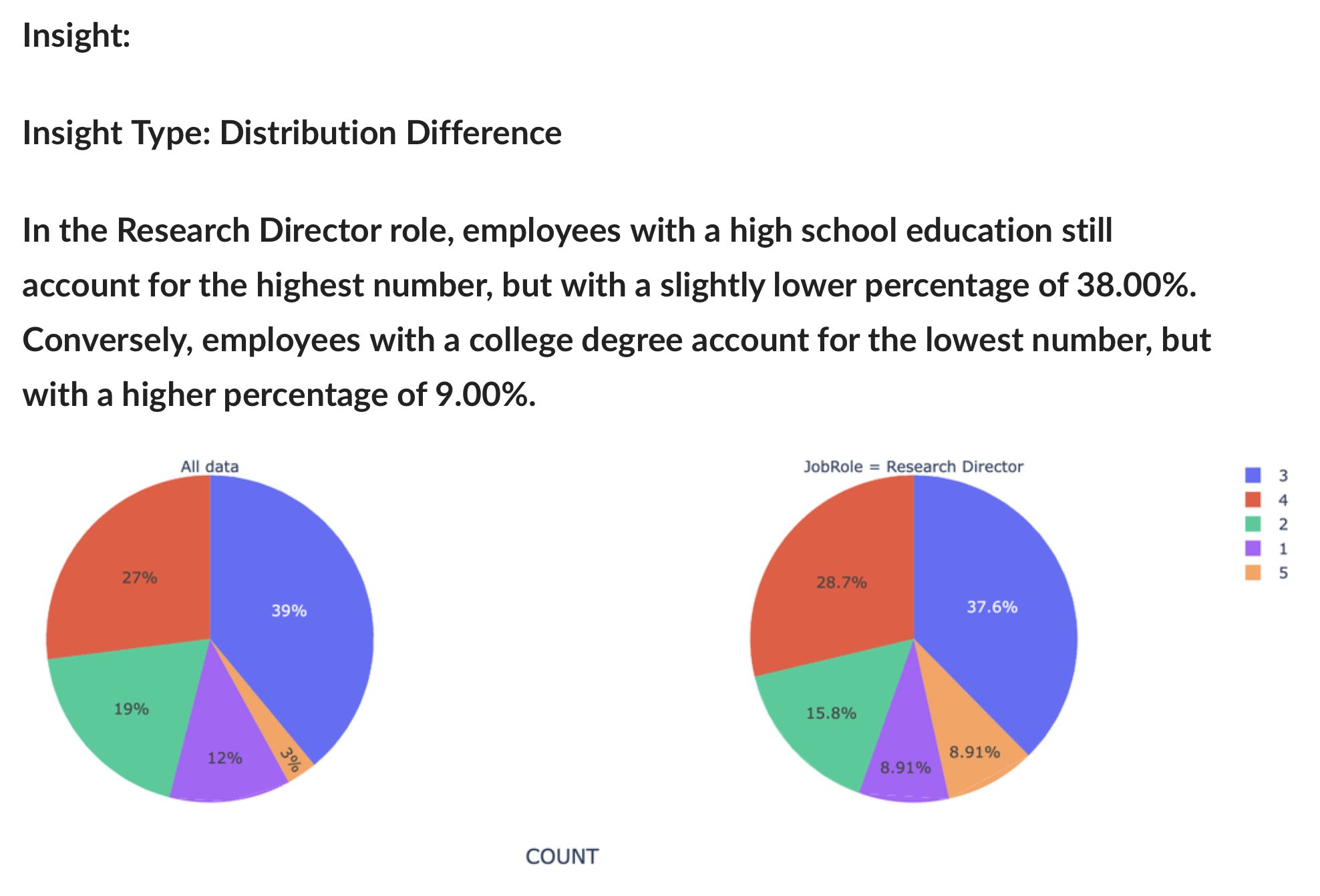} \\
      \vspace{10pt}
    \end{minipage} \\
\cline{2-3}

& \multirow{3}{4.5cm}{What is the distribution of Attrition rates across different Departments?} &
 \begin{minipage}{6cm}
      \centering
      \vspace{10pt}
      \includegraphics[width=6cm]{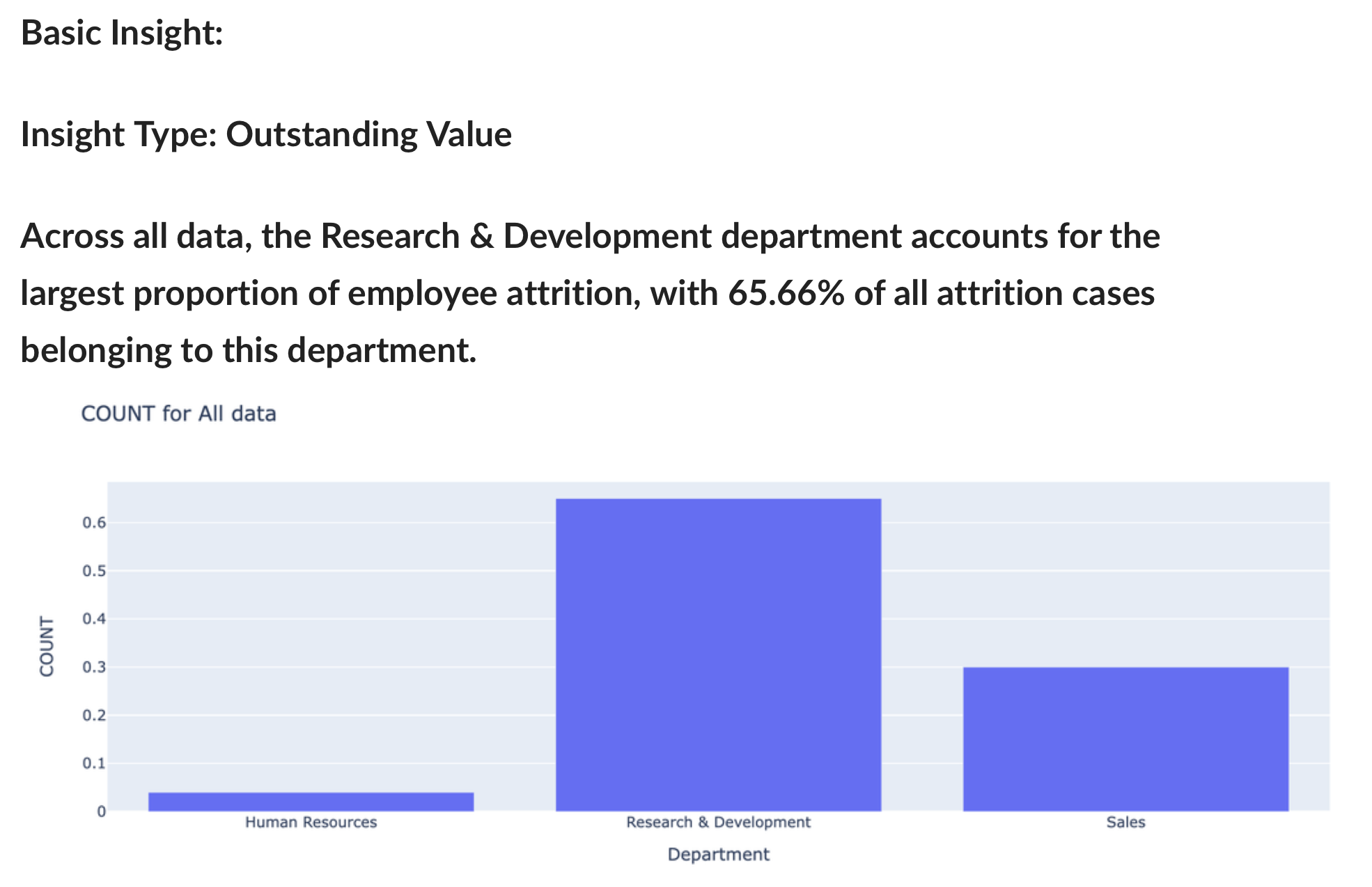} \\
    \end{minipage} \\

 &  &
 \begin{minipage}{6cm}
      \centering
      \vspace{10pt}
      \includegraphics[width=6cm]{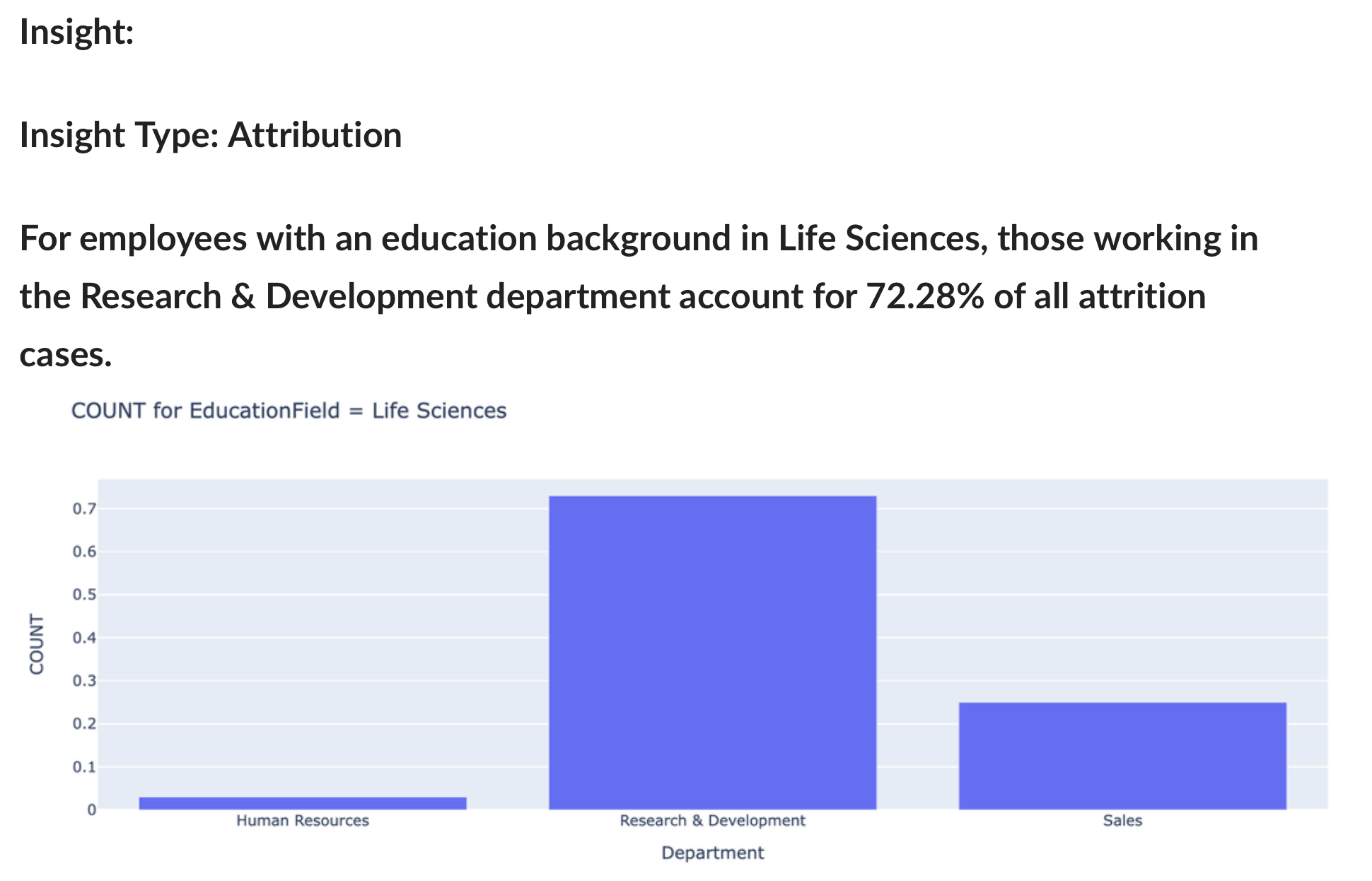} \\
      \vspace{10pt}
    \end{minipage} \\

 \hline

\end{tabular}
\caption{ QUIS Results for Employee Attrition Dataset}
\end{center}
\end{table*}


\begin{figure*}[ht]
\centering
  \makebox[\textwidth]{\includegraphics[width=1.2\textwidth]{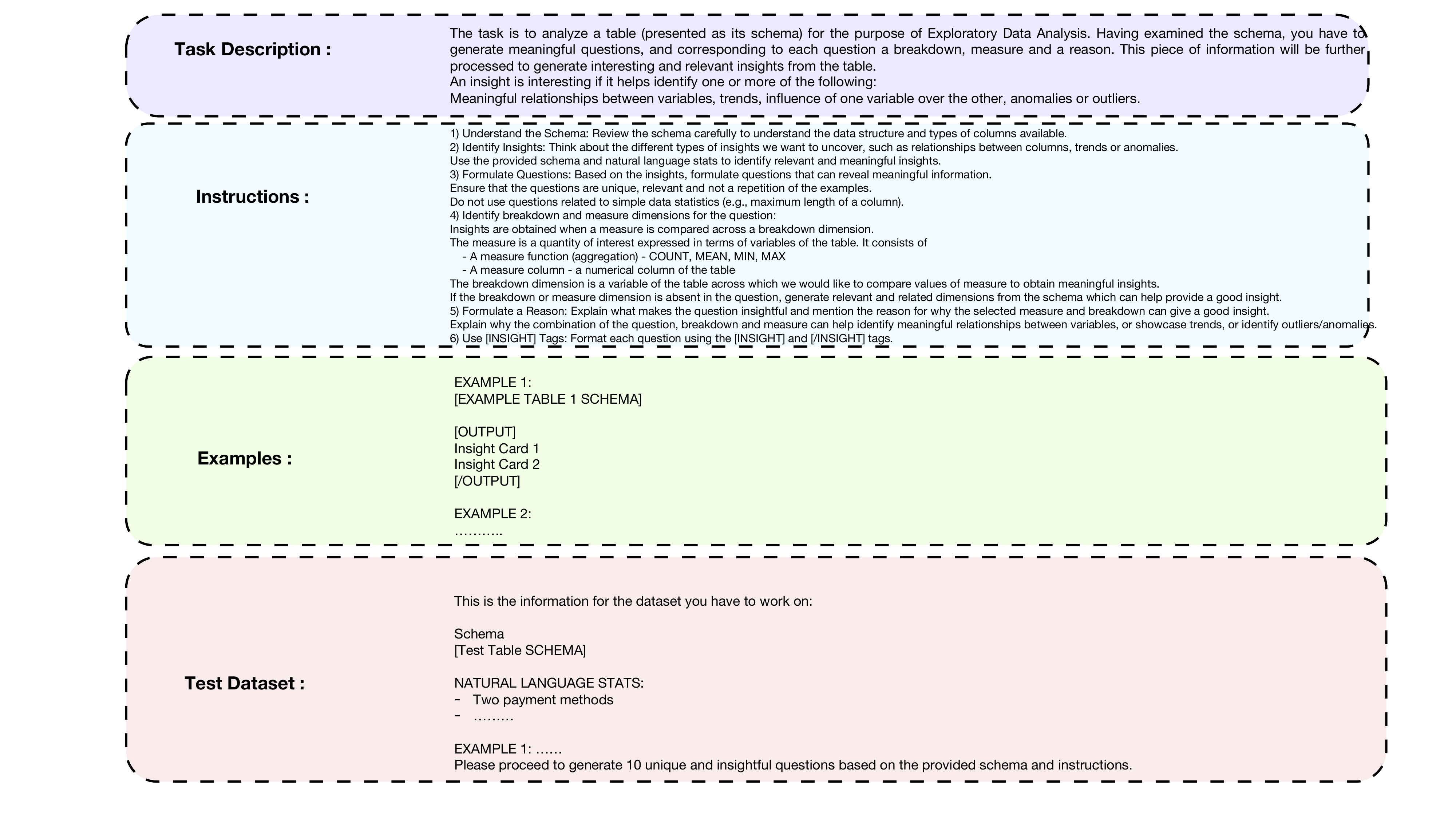}}
  \caption{QUGen Prompt Template}
  \label{fig:QUGen_prompt}
\end{figure*}

\begin{figure*}[ht]
\centering
  \includegraphics[width=1.1\textwidth]{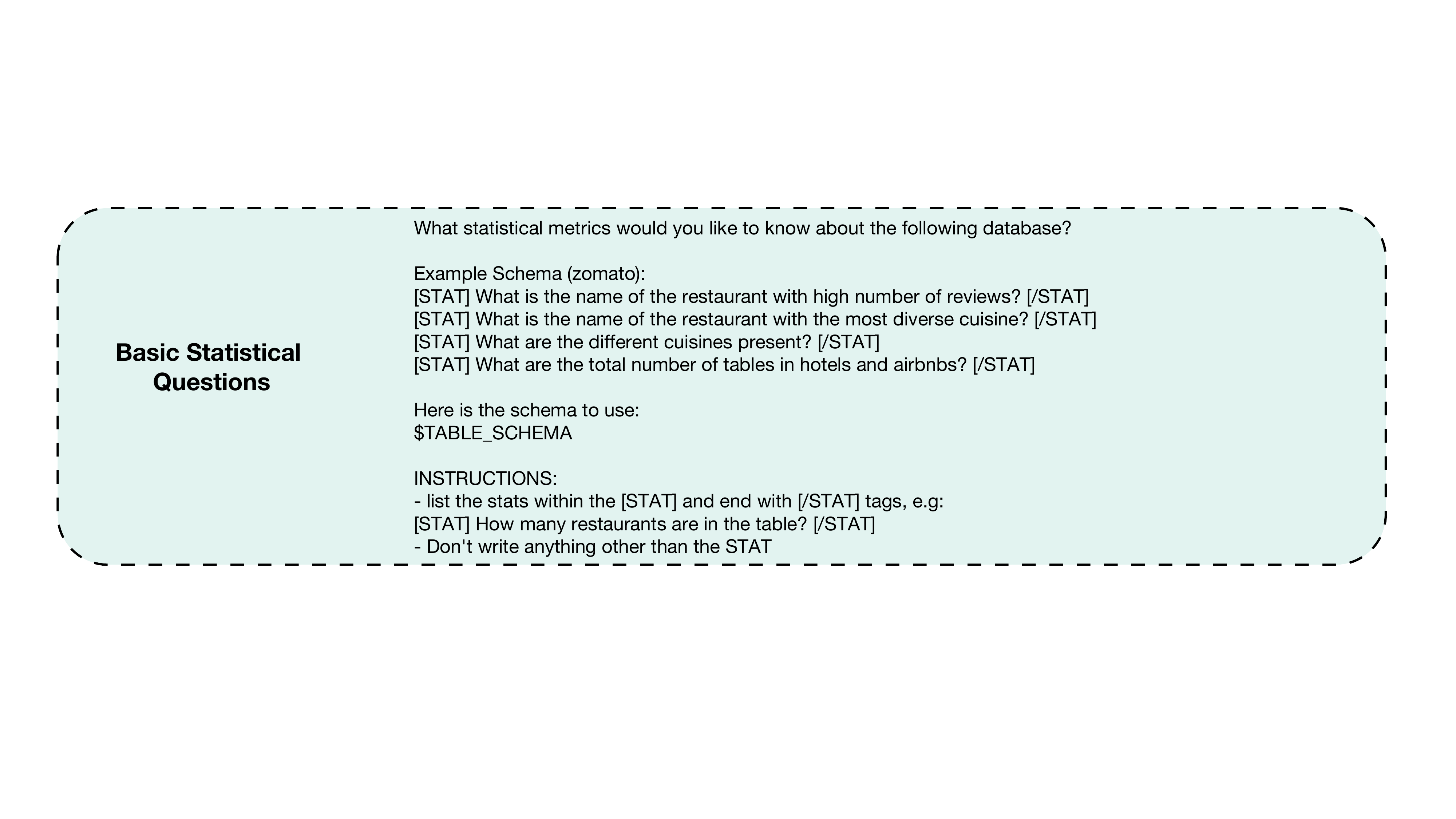}
  \caption{Natural Language Statistics Prompt Template.}
  \label{fig:prompt_template}
\end{figure*}

\end{document}